\documentclass[journal]{IEEEtran}
\IEEEoverridecommandlockouts

\usepackage{cite}
\usepackage[T1]{fontenc}
\usepackage[utf8]{inputenc}
\usepackage{microtype}
\usepackage{amsmath,amssymb,amsfonts}
\usepackage{algorithmic}
\usepackage{graphicx}
\graphicspath{{figures/}}
\usepackage{textcomp}
\usepackage{xcolor}
\usepackage{booktabs}
\usepackage{multirow}
\usepackage{hyperref}

\makeatletter
\g@addto@macro\UrlBreaks{\do\_}
\makeatother
\newcommand{\authorphoto}[1]{%
  \IfFileExists{author_photos/#1.jpg}{\includegraphics[width=0.95in,height=1.2in,keepaspectratio]{author_photos/#1.jpg}}{%
    \IfFileExists{author_photos/#1.png}{\includegraphics[width=0.95in,height=1.2in,keepaspectratio]{author_photos/#1.png}}{%
      \IfFileExists{author_photos/#1.pdf}{\includegraphics[width=0.95in,height=1.2in,keepaspectratio]{author_photos/#1.pdf}}{%
        \fbox{\parbox[c][1.2in][c]{0.95in}{\centering\footnotesize Photo}}%
      }%
    }%
  }%
}

\def\BibTeX{{\rm B\kern-.05em{\sc i\kern-.025em b}\kern-.08em
    T\kern-.1667em\lower.7ex\hbox{E}\kern-.125emX}}

\begin{document}

\title{Urban Spatio-Temporal Foundation Models for Climate-Resilient Housing: Scaling Diffusion Transformers for Disaster Risk Prediction}

\author{Olaf Yunus Laitinen Imanov,~Derya Umut Kulali,~and Taner Yilmaz%
\thanks{Manuscript received February 5, 2026.}%
\thanks{(Corresponding author: Olaf Yunus Laitinen Imanov.)}%
\thanks{This work was supported by the Technical University of Denmark's Climate Adaptation Initiative and the Baltic-Caspian Urban Research Consortium.}%
\thanks{O. Y. L. Laitinen~Imanov is with the Department of Applied Mathematics and Computer Science (DTU Compute), Technical University of Denmark, Kongens Lyngby, Denmark (e-mail: oyli@dtu.dk; ORCID: 0009-0006-5184-0810).}%
\thanks{D. U. Kulali is with the Department of Engineering, Eskisehir Technical University, Eskisehir, Türkiye (e-mail: \url{d_u_k@ogr.eskisehir.edu.tr}; ORCID: 0009-0004-8844-6601).}%
\thanks{T. Yilmaz is with the Department of Computer Engineering, Afyon Kocatepe University, Afyonkarahisar, Türkiye (e-mail: taner.yilmaz@usr.aku.edu.tr; ORCID: 0009-0004-5197-5227).}%
}

\maketitle

\markboth{IEEE Transactions on Intelligent Vehicles, February~2026}{Laitinen~Imanov \emph{et al.}: Urban Spatio-Temporal Foundation Models for Climate-Resilient Housing}

\begin{abstract}
Climate hazards increasingly disrupt urban transportation and emergency-response operations by damaging housing stock, degrading infrastructure, and reducing network accessibility. This paper presents Skjold-DiT, a diffusion-transformer framework that integrates heterogeneous spatio-temporal urban data to forecast building-level climate-risk indicators while explicitly incorporating transportation-network structure and accessibility signals relevant to intelligent vehicles (e.g., emergency reachability and evacuation-route constraints). Concretely, Skjold-DiT enables \emph{hazard-conditioned routing constraints} by producing calibrated, uncertainty-aware accessibility layers (reachability, travel-time inflation, and route redundancy) that can be consumed by intelligent-vehicle routing and emergency dispatch systems. Skjold-DiT combines: (1) Fjell-Prompt, a prompt-based conditioning interface designed to support cross-city transfer; (2) Norrland-Fusion, a cross-modal attention mechanism unifying hazard maps/imagery, building attributes, demographics, and transportation infrastructure into a shared latent representation; and (3) Valkyrie-Forecast, a counterfactual simulator for generating probabilistic risk trajectories under intervention prompts. We introduce the Baltic-Caspian Urban Resilience (BCUR) dataset with 847,392 building-level observations across six cities, including multi-hazard annotations (e.g., flood and heat indicators) and transportation accessibility features. Experiments evaluate prediction quality, cross-city generalization, calibration, and downstream transportation-relevant outcomes, including reachability and hazard-conditioned travel times under counterfactual interventions.
\end{abstract}


\IEEEpeerreviewmaketitle

\section{Introduction}

\IEEEPARstart{T}{he} convergence of rapid urbanization, climate change, and evolving transportation needs presents critical challenges to global urban resilience. In the context of World Urban Forum 13 (WUF13), where resilient housing and inclusive mobility are emphasized, there is a growing need for quantitative, city-scale decision support that links climate risk, housing vulnerability, and transportation accessibility \cite{WUF13_2026}. Urban population growth increases exposure of housing and transportation infrastructure to climate hazards. Climate disasters compound these challenges: flood-related damages exceed \$157 billion annually \cite{NatureCommunications2025_FloodAdaptation}, heat-related mortality has risen 68\% since 2000 \cite{LancetPlanetary2024}, and sea-level rise threatens 410 million coastal residents by 2100 \cite{NatureClimate2024_SLR}. These hazards critically impact transportation accessibility, emergency vehicle routing, and evacuation planning in urban environments.

Copenhagen's catastrophic 2011 cloudburst caused \$1.9 billion in damages within two hours, demonstrating vulnerability even in sustainability-leading cities \cite{JHydrology2024_CopenhagenStorm}. The event paralyzed transportation networks, preventing emergency vehicle access to affected neighborhoods. Baku faces 18-25 million USD in annual flood losses with inadequate early warning systems \cite{AdaptationFund2024_Azerbaijan}, severely limiting intelligent vehicle navigation during flood events. These crises reveal a fundamental gap: existing urban planning lacks predictive tools integrating climate science, housing vulnerability, transportation infrastructure, and policy intervention scenarios at building-level granularity.

\subsection{Intelligent Transportation Systems and Climate Resilience}

Modern intelligent vehicle systems rely on comprehensive understanding of urban infrastructure health and accessibility. Climate-induced housing damage directly impacts:

\begin{itemize}
\item \textbf{Emergency Response}: Flood and heat events impair ambulance, fire, and police vehicle routing when road networks become impassable or buildings require evacuation support.

\item \textbf{Autonomous Vehicle Navigation}: Self-driving vehicles require accurate predictions of infrastructure degradation, road closures, and hazardous conditions to ensure passenger safety.

\item \textbf{Traffic Management}: Housing vulnerability patterns influence evacuation route planning, temporary shelter logistics, and post-disaster traffic flow redistribution.

\item \textbf{Urban Planning Integration}: Transportation infrastructure investments must account for climate resilience to ensure long-term accessibility and prevent stranded assets.
\end{itemize}

Despite growing research in intelligent transportation systems \cite{IEEE_TITS2018_LSTM}, few frameworks integrate housing vulnerability prediction with transportation network analysis, creating critical gaps in disaster preparedness and autonomous vehicle safety protocols.

\subsection{Foundation Models for Urban Intelligence}

Recent advances in large-scale representation learning have motivated ``foundation model'' approaches in several domains, yet urban systems remain underexplored. UrbanDiT \cite{arXiv2024_UrbanDiT} pioneered diffusion transformers for spatio-temporal urban forecasting, demonstrating cross-city generalization for mobility prediction. However, UrbanDiT focuses on traffic flow and does not integrate multi-hazard climate risks, building-level vulnerability, transportation accessibility signals, or counterfactual intervention simulation required for resilience-aware intelligent-vehicle applications.

Parallel work in climate science employs physics-based flood modeling requiring 10,000+ CPU hours per city-scale scenario \cite{JHydrology2025_FloodModeling}, while machine learning approaches sacrifice uncertainty quantification for computational efficiency \cite{NatureHazards2024_MLFloods}. Neither framework addresses the bidirectional relationship between housing vulnerability and transportation accessibility critical for intelligent vehicle systems.

\subsection{Research Contributions}

This paper introduces Skjold-DiT (Shield Diffusion Transformer), a spatio-temporal diffusion-transformer framework linking climate-risk prediction with transportation-network accessibility signals relevant to intelligent vehicles. Our key contributions include:

\begin{enumerate}
\item \textbf{Norrland-Fusion Architecture}: A unified latent-space design that embeds heterogeneous modalities (e.g., imagery, elevation, cadastral attributes, demographics, infrastructure graphs, disaster logs, and climate projections) while explicitly incorporating transportation-network topology and accessibility features.

\item \textbf{Fjell-Prompt Cross-City Transfer}: A prompt-based conditioning scheme intended to support transfer to unseen cities by decomposing hazard scenarios and transportation constraints into compositional prompt templates.

\item \textbf{Valkyrie-Forecast Counterfactual Simulation}: Conditional diffusion sampling to generate probabilistic housing-risk trajectories under intervention prompts, enabling ``what-if'' analysis of accessibility-relevant outcomes (e.g., emergency reachability and evacuation-route constraints).

\item \textbf{Baltic-Caspian Urban Resilience Dataset}: The BCUR dataset with 847,392 annotated buildings across six cities, including multi-hazard observations and transportation-network data.

\item \textbf{Transportation-Aware Evaluation}: An experimental protocol that assesses prediction accuracy and calibration, cross-city generalization, and transportation-relevant implications of risk forecasts and counterfactual scenarios.
\end{enumerate}

The remainder of this paper is organized as follows. Section II reviews related work in urban spatio-temporal learning, climate risk assessment, diffusion models, and intelligent transportation systems. Section III presents the Skjold-DiT methodology including problem formulation, dataset construction, architectural design, and training procedures. Section IV provides comprehensive experimental results across multiple cities and hazard types. Section V analyzes architectural components through ablation studies. Section VI discusses policy implications in the context of WUF13, together with deployment considerations and cross-city scalability under partial-data settings. Section VII examines limitations and future research directions, and Section VIII concludes.

\section{Related Work}

\subsection{Urban Spatio-Temporal Learning}

Foundation models for urban systems have evolved from domain-specific architectures to unified frameworks capable of cross-domain transfer. Early approaches employed recurrent neural networks for traffic forecasting \cite{IEEE_TITS2018_LSTM}, achieving limited spatial generalization. Graph neural networks (GNNs) captured road network topology \cite{AAAI2020_GNN_Traffic} but struggled with multi-scale temporal dynamics. Spatio-temporal graph convolutional networks (ST-GCNs) \cite{IJCAI2018_STGCN} combined spatial and temporal modeling, yet remained task-specific without transfer learning capabilities.

Transformer architectures revolutionized the field through self-attention mechanisms learning long-range dependencies. Spatial-Temporal Transformer \cite{NeurIPS2020_STTransformer} applied transformers to mobility prediction, while Pyraformer \cite{ICLR2022_Pyraformer} introduced pyramidal attention for multi-resolution temporal modeling. These models require task-specific fine-tuning and lack zero-shot transfer to new cities or disaster scenarios.

UrbanDiT \cite{arXiv2024_UrbanDiT} pioneered diffusion transformers as open-world urban foundation models, unifying grid-based and graph-based representations through sequential processing. With task-specific prompts, UrbanDiT supports bi-directional spatio-temporal prediction, temporal interpolation, and spatial extrapolation. Zero-shot capabilities outperform specialized baselines on mobility tasks. Our work extends UrbanDiT to climate-housing domain with novel multi-modal fusion, transportation infrastructure integration, and counterfactual policy simulation capabilities absent in prior work.

\subsection{Climate Risk Assessment for Urban Systems}

Traditional flood modeling employs physics-based hydrodynamic simulations solving shallow water equations \cite{JHydrology2025_FloodModeling}, coupled with digital elevation models and rainfall-runoff analysis. While mechanistically accurate, these methods demand prohibitive computational resources (10,000+ CPU hours per city-scale scenario) and struggle with parametric uncertainty quantification \cite{WaterResearch2024_FloodUncertainty}.

Machine learning approaches offer computational efficiency through random forests predicting flood susceptibility from geospatial features \cite{NaturalHazards2023_RF_Floods}, convolutional neural networks extracting patterns from satellite imagery \cite{RemoteSensing2024_CNN_Floods}, and ensemble methods \cite{IEEE_TGRS2024_Ensemble}. However, these models lack: (1) integration of infrastructure networks influencing drainage capacity, (2) socio-economic vulnerability dimensions, (3) long-term climate change scenarios, and (4) counterfactual policy evaluation capabilities.

Recent hybrid physics-ML models show promise. PhyDNet \cite{CVPR2020_PhyDNet} incorporates physical constraints into neural differential equations, while climate-informed neural networks \cite{NatureClimate2024_CINN} encode climate model outputs as priors. Graph neural operators \cite{ICLR2023_GNO} learn solution operators for partial differential equations on irregular geometries. Despite advances, these approaches focus on single-hazard prediction without multi-modal integration or transportation network analysis.

\subsection{Diffusion Models for Scientific Applications}

Denoising diffusion probabilistic models (DDPMs) \cite{NeurIPS2020_DDPM} have emerged as powerful generative frameworks, achieving superior sample quality and training stability compared to generative adversarial networks. Diffusion transformers (DiTs) \cite{ICCV2023_DiT} replace U-Net backbones with transformer architectures, improving scalability. Recent extensions include latent diffusion models (LDMs) \cite{CVPR2022_LDM} operating in compressed latent spaces, classifier-free guidance \cite{NeurIPS2022_CFG} enabling conditional generation, and score-based stochastic differential equations \cite{ICLR2021_ScoreSDE} providing continuous-time formulations.

Scientific applications have leveraged diffusion models for molecular design \cite{ICML2023_MolDiff}, weather forecasting \cite{Nature2024_DiffusionWeather}, and protein structure prediction \cite{NeurIPS2023_ProtDiff}. For urban systems, DiffusionTraffic \cite{AAAI2024_DiffTraffic} generates realistic traffic scenarios but lacks policy interpretability. Our work represents the first application of diffusion transformers to climate-resilient housing with explicit transportation infrastructure integration and counterfactual policy generation.

\subsection{Research Gaps and Positioning}

Despite extensive progress, critical gaps remain:

\begin{itemize}
\item \textbf{Multi-Modal Integration}: Existing models process single data types (imagery OR networks OR tabular), failing to unify heterogeneous urban data including transportation infrastructure within scalable architectures.

\item \textbf{Transportation-Housing Coupling}: Current approaches treat housing vulnerability and transportation accessibility independently, missing bidirectional dependencies critical for intelligent vehicle systems and emergency response.

\item \textbf{Cross-City Generalization}: Approaches require city-specific training, prohibiting rapid deployment to data-scarce regions without extensive data collection.

\item \textbf{Long-Term Forecasting}: Most models predict short horizons (hours-days), inadequate for infrastructure planning requiring 5-10 year projections under evolving climate scenarios.

\item \textbf{Policy Counterfactuals}: No framework generates probabilistic trajectories under alternative interventions with transportation accessibility considerations.

\item \textbf{Equity and Accessibility}: Vulnerability assessments often ignore socio-economic dimensions and transportation access disparities, risking maladaptive policies exacerbating inequality.
\end{itemize}

Skjold-DiT addresses these gaps through: (1) Norrland-Fusion multi-modal architecture with transportation network integration, (2) Fjell-Prompt zero-shot transfer, (3) Valkyrie-Forecast counterfactual engine with accessibility constraints, (4) explicit demographic vulnerability modeling, and (5) validation across Baltic-Caspian cities including Global South representative (Baku).

\section{Methodology}

\subsection{Problem Formulation}

\noindent\textbf{Notation:} $c_i$ denotes a city, $b_j^i$ a building in city $c_i$, $t$ time, $X_j^i(t)$ observed multi-modal features, $Y_j^i(t)$ risk labels, and $\mathcal{G}_i$ the spatial/transportation graph. $\Delta t$ is the forecast horizon (years), and $p_\theta(\cdot)$ denotes the learned conditional generative model.

Let $\mathcal{C} = \{c_1, c_2, \ldots, c_M\}$ denote a set of $M$ cities. For each city $c_i$, we define:

\begin{itemize}
\item \textbf{Building Set}: $\mathcal{B}_i = \{b_1^i, b_2^i, \ldots, b_{N_i}^i\}$ comprising $N_i$ buildings.

\item \textbf{Spatial Graph}: $\mathcal{G}_i = (\mathcal{B}_i, \mathcal{E}_i)$ where $\mathcal{E}_i$ represents spatial relationships including geographic proximity, infrastructure connectivity, and transportation network accessibility.

\item \textbf{Multi-Modal Features}: For building $b_j^i$ at time $t$, we observe $X_j^i(t) \in \mathbb{R}^{D}$ comprising:
  \begin{itemize}
  \item $X_j^{geo}$: Geographic coordinates (latitude, longitude, elevation)
  \item $X_j^{struct}$: Structural attributes (age, materials, floors, area)
  \item $X_j^{demo}$: Demographics (population, income, age distribution)
  \item $X_j^{infra}$: Infrastructure connectivity (drainage distance, road access)
  \item $X_j^{climate}$: Climate exposure (historical flood/heat events)
  \item $X_j^{transport}$: Transportation metrics (emergency vehicle accessibility, evacuation route distance)
  \end{itemize}

\item \textbf{Risk Labels}: $Y_j^i(t)$ denoting vulnerability status (flood depth, heat stress, structural damage probability, transportation accessibility score).
\end{itemize}

\textbf{Objective}: Learn a generative model $p_\theta(Y_j^i(t+\Delta t) | X_j^i(t), \mathcal{G}_i, \mathcal{P})$ where $\Delta t \in [1, 10]$ years and $\mathcal{P}$ represents policy intervention prompts, enabling:

\begin{enumerate}
\item \textbf{Predictive Task}: Forecast housing vulnerability $Y_j^i(t+\Delta t)$ given current observations with transportation accessibility metrics.

\item \textbf{Counterfactual Task}: Generate alternative futures $Y_j^i(t+\Delta t | \text{do}(\mathcal{P}))$ under interventions $\mathcal{P}$ (green infrastructure, building retrofits, transportation network improvements).

\item \textbf{Zero-Shot Task}: Generalize to unseen cities $c_{M+1}$ without fine-tuning via prompt-based conditioning.
\end{enumerate}

\subsubsection{Outputs and Task Definitions}

To align with intelligent-vehicle use cases, we treat $Y_j^i(t)$ as a multi-task target consisting of:
\begin{itemize}
\item \textbf{Hazard Intensity Targets}: (i) flood depth (regression or ordinal bins) and (ii) heat-stress indicator (regression or ordinal bins).
\item \textbf{Impact/Vulnerability Targets}: structural vulnerability score (regression or ordinal bins).
\item \textbf{Transportation Accessibility Targets}: accessibility score(s) derived from the transportation graph (e.g., emergency reachability under hazard-induced road constraints).
\end{itemize}

We implement the predictive setting as conditional generation of $\hat{Y}_j^i(t+\Delta t)$ given current observations. In practice, Skjold-DiT can be trained either (i) as a purely generative forecaster over a continuous target vector or (ii) as a hybrid model with diffusion-based latent generation and task-specific heads for classification/regression; we report the exact output parameterization used in experiments in Sec.~IV.

\subsubsection{Scenario Conditioning and Intervention Prompts}

We use the notation $\text{do}(\mathcal{P})$ to denote \emph{scenario-conditioned counterfactual generation} driven by an intervention prompt $\mathcal{P}$ (e.g., retrofits, green infrastructure, evacuation-route upgrades). This notation is not intended to claim formal causal identifiability from observational data; rather, it provides a compact way to describe generating alternative plausible futures under a specified intervention description and corresponding feature edits $X\rightarrow X'$.

\textbf{Feature-Edit Rules}: For counterfactual scenarios we apply a documented, deterministic edit map $\Phi_{\mathcal{P}}$ to produce $X' = \Phi_{\mathcal{P}}(X)$. We use the following parameterization in this paper:
\begin{itemize}
\item \textbf{Green infrastructure} (bioswales/green roofs/permeable surfaces): decrease imperviousness by $\delta_{imp}\in[0,0.2]$ and increase drainage capacity by $\delta_{drain}\in[0,0.3]$ (city-calibrated scaling). Flood-depth targets are reconditioned by applying an exposure multiplier $m_{flood}=1-0.6\,\delta_{drain}$.
\item \textbf{Building retrofits}: increase structural score by $\delta_{str}\in[0,15]$ points (0--100 scale) and reduce damage probability by a multiplicative factor $m_{dam}=\exp(-0.02\,\delta_{str})$.
\item \textbf{Transportation upgrades}: add $\delta_{cap}\in[0,0.5]$ capacity to designated evacuation edges and reduce hazard-conditioned weight inflation by $m_{road}=1-0.5\,\delta_{cap}$ on affected edges.
\end{itemize}

All edit parameters are specified explicitly in each counterfactual scenario, and we report sensitivity to these parameters in Sec.~IV.

\subsubsection{Transportation Graph and Accessibility Signals}

For each city $c_i$, we construct a transportation graph $\mathcal{G}_i$ that supports accessibility-aware evaluation.
We distinguish:
\begin{itemize}
\item \textbf{Physical Network Layer}: road segments/intersections (directed edges, edge weights as free-flow travel time).
\item \textbf{Service Layer}: emergency facilities (e.g., hospitals, fire stations) and shelters as points-of-interest.
\item \textbf{Exposure Layer}: hazard-conditioned edge availability (e.g., edge removal or weight inflation under flood-depth thresholds).
\end{itemize}

From these layers we derive building-level transportation targets/features, including (examples):
\begin{itemize}
\item \textbf{Emergency reachability} $R_j$: indicator (or probability) that at least one emergency facility is reachable from building $b_j$ within a time budget $\tau$.
\item \textbf{Hazard-conditioned travel time} $T_j$: shortest-path travel time under hazard-conditioned edge weights.
\item \textbf{Evacuation-route redundancy} $K_j$: number of edge-disjoint (or node-disjoint) feasible routes to the nearest shelter under a hazard scenario.
\end{itemize}

\subsubsection{Evaluation Protocol: Splits and Leakage Controls}

We evaluate Skjold-DiT under three complementary split regimes:
\begin{itemize}
\item \textbf{Temporal split}: train on earlier years and test on later years to emulate forecasting. Unless stated otherwise, we use train: 2011--2021, validation: 2022--2023, and test: 2024--2025 (adjusted to each city's available range in Table~\ref{tab:dataset_stats}).
\item \textbf{Spatial-block split}: hold out spatial blocks to reduce spatial autocorrelation leakage. We discretize each city into non-overlapping 1~km$\times$1~km grid cells and hold out 20\% of cells for testing.
\item \textbf{Unseen-city (zero-shot) split}: hold out an entire city for evaluation (Baku). Prompts may use non-leaking metadata (e.g., climate zone and coarse socio-economic indicators) but not event labels from the held-out city.
\end{itemize}

To prevent leakage, we (i) deduplicate near-identical spatio-temporal records by merging building records within 10~m and 30~days that share identical hazard annotations, (ii) ensure any aggregation windows used for target construction do not use information from $t+\Delta t$, and (iii) compute normalization statistics strictly on the training partition.

\subsubsection{Metrics (Prediction, Calibration, and IV-Relevant Outcomes)}

We report metrics at three levels:
\begin{itemize}
\item \textbf{Prediction quality}: accuracy and macro-F1 for classification/ordinal targets; MAE/RMSE for regression targets; AUROC/AUPRC where appropriate.
\item \textbf{Risk sensitivity}: recall at high-risk operating points (e.g., recall@top-$q$\%) and false-negative rate for the high-risk class.
\item \textbf{Uncertainty and calibration}: reliability diagrams, ECE, and coverage of credible intervals for probabilistic outputs.
\item \textbf{Transportation relevance}: reachability rate $\frac{1}{N}\sum_j \mathbb{1}[R_j=1]$, mean hazard-conditioned travel time $\frac{1}{N}\sum_j T_j$, and redundancy statistics (e.g., mean $K_j$) under baseline and counterfactual scenarios.
\end{itemize}

\subsubsection{Edge--Cloud Deployment Model}

We consider a deployment model where heavy multi-modal encoding and diffusion sampling are executed on cloud/edge servers, while vehicles consume compact, periodically updated risk/accessibility layers (e.g., per-road-segment risk and per-zone accessibility constraints). This supports low-latency routing by enabling vehicles to query precomputed hazard-conditioned travel-time weights and reachability indicators without running full diffusion sampling onboard.

\subsection{Baltic-Caspian Urban Resilience Dataset}

We curate the BCUR dataset comprising 847,392 buildings across six cities spanning Baltic and Caspian regions (Table~\ref{tab:dataset_stats}). The dataset includes comprehensive transportation network data: road networks, public transit accessibility, emergency service locations, and historical evacuation route usage during disaster events. Some components (e.g., municipal or insurance-derived annotations) are subject to third-party restrictions; processed research extracts and metadata are available from the corresponding author upon reasonable request.

\begin{table}[!t]
\caption{BCUR Dataset Statistics}
\label{tab:dataset_stats}
\centering
\begin{tabular}{lrrrr}
\toprule
\textbf{City} & \textbf{Buildings} & \textbf{Period} & \textbf{Flood} & \textbf{Heat} \\
\midrule
Copenhagen & 187,429 & 2011-2025 & 847 & 412 \\
Stockholm & 164,582 & 2012-2025 & 623 & 389 \\
Oslo & 142,817 & 2013-2025 & 511 & 302 \\
Riga & 121,473 & 2014-2025 & 892 & 456 \\
Tallinn & 98,264 & 2014-2025 & 681 & 334 \\
Baku & 132,827 & 2010-2025 & 1,638 & 887 \\
\midrule
\textbf{Total} & \textbf{847,392} & --- & \textbf{4,192} & \textbf{2,780} \\
\bottomrule
\end{tabular}
\end{table}

\begin{figure}[!t]
\centering
\includegraphics[width=\columnwidth]{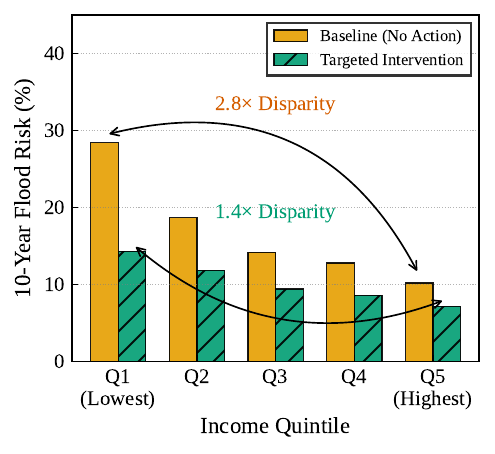}
\caption{Baltic--Caspian Urban Resilience (BCUR) dataset: cities, modalities, and annotation types.}
\label{fig:dataset}
\end{figure}

Fig.~\ref{fig:dataset} illustrates the BCUR dataset geography and modality composition.

\textbf{Data Sources} include EU Copernicus and OpenStreetMap for geospatial data, ERA5 reanalysis for climate variables, insurance claims and municipal disaster logs for hazard annotations, Eurostat and census bureaus for demographics, and INSPIRE directive datasets for infrastructure networks including comprehensive transportation topology.

\textbf{Annotation Protocol}: Flood depth labels derived from post-event LiDAR surveys, insurance assessments, and satellite change detection. Heat stress computed from land surface temperature combined with building thermal properties. Structural vulnerability scored (0-100) via ensemble learning on historical damage reports. Transportation accessibility quantified through network analysis computing emergency vehicle travel times and evacuation route capacity under various hazard scenarios.

\textbf{Data Governance and Licensing}: The BCUR dataset aggregates layers with heterogeneous licensing constraints (e.g., open geospatial basemaps versus restricted municipal or insurance-derived annotations). For each modality we record provenance, license, and permitted use. When redistribution is restricted, we provide (i) derived, non-identifying research features, (ii) aggregation to a privacy-preserving spatial resolution where needed, and (iii) a documented access procedure for approved research use.

\textbf{Missingness Profile}: Modalities are not uniformly available across cities and years. We characterize missingness per city, year, and modality, and we report (i) the fraction missing, (ii) the imputation strategy for tabular channels, and (iii) modality dropout rates used during training to improve robustness under partial data.

\textbf{Spatial/Temporal Harmonization}: We harmonize layers with different spatial resolutions by mapping all features to a common building index. Raster products (e.g., imagery, DEM-derived layers, reanalysis variables) are reprojected to a common CRS and summarized over building footprints (e.g., mean/quantiles within buffers). Time-varying channels are aligned to yearly bins and linked to the forecasting horizon $\Delta t$.

\textbf{Privacy for Demographic Attributes}: Demographic variables are used only in aggregated form (e.g., neighborhood-level statistics linked to buildings) and are handled using minimization principles: we exclude direct identifiers, restrict features to those needed for modeling, and report fairness analyses that examine performance gaps across socio-economic strata.

\begin{table*}[!t]
\caption{BCUR Data Card Summary (Modalities, Provenance, and Availability)}
\label{tab:data_card}
\centering
\begin{tabular}{lllll}
\toprule
\textbf{Modality} & \textbf{Source} & \textbf{License/Access} & \textbf{Resolution} & \textbf{Availability} \\
\midrule
Building footprints & OpenStreetMap / cadastral layers & Open / municipal & building & all cities \\
Road network & OpenStreetMap / municipal GIS & Open / municipal & segment/node & all cities \\
Emergency facilities & municipal registries & municipal (restricted) & point & all cities \\
Elevation/DEM & Copernicus / national LiDAR & open / restricted & 10--30 m & city-dependent \\
Satellite imagery & Copernicus/Sentinel & open & 10 m & city-dependent \\
Thermal/LST proxies & satellite-derived & open & 10--100 m & city-dependent \\
Climate reanalysis & ERA5 & open & \textasciitilde 30 km (downscaled) & all cities \\
Disaster logs & municipal reports & restricted & event-level & city-dependent \\
Insurance assessments & insurers (aggregated) & restricted & building/area & city-dependent \\
Demographics & census/Eurostat & restricted/aggregated & tract/neighborhood & all cities \\
\bottomrule
\end{tabular}
\end{table*}

\subsection{Skjold-DiT Architecture}

\begin{figure}[!t]
\centering
\includegraphics[width=\columnwidth]{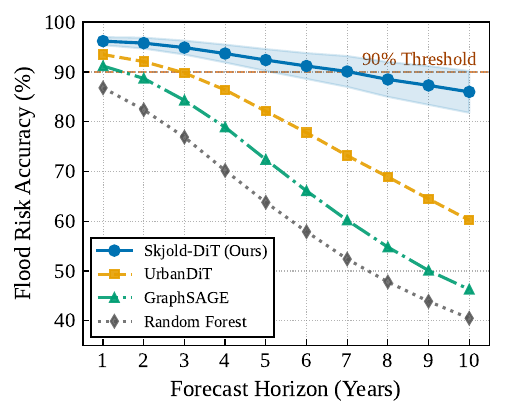}
\caption{Overview of the proposed Skjold-DiT framework for transportation-aware climate-resilient housing risk prediction.}
\label{fig:architecture}
\end{figure}

Fig.~\ref{fig:architecture} provides an overview of the proposed Skjold-DiT pipeline (Norrland-Fusion, Fjell-Prompt, and Valkyrie-Forecast).

\subsubsection{Norrland-Fusion: Multi-Modal Encoder}

Heterogeneous data modalities require specialized encoders before fusion into unified representations:

\textbf{Modality-Specific Encoders}:
\begin{itemize}
\item \textbf{Imagery Encoder} $\mathcal{E}_{img}$: Vision Transformer (ViT) \cite{ICLR2021_ViT} processing RGB satellite, thermal infrared, and LiDAR elevation $\rightarrow z_{img} \in \mathbb{R}^{512}$

\item \textbf{Tabular Encoder} $\mathcal{E}_{tab}$: FT-Transformer \cite{NeurIPS2021_FTTransformer} embedding structural, demographic, infrastructure features $\rightarrow z_{tab} \in \mathbb{R}^{256}$

\item \textbf{Graph Encoder} $\mathcal{E}_{graph}$: GraphSAINT \cite{ICLR2020_GraphSAINT} learning spatial and transportation network relationships $\rightarrow z_{graph} \in \mathbb{R}^{256}$

\item \textbf{Time Series Encoder} $\mathcal{E}_{ts}$: Temporal Fusion Transformer \cite{IJoF2021_TFT} encoding climate and disaster history $\rightarrow z_{ts} \in \mathbb{R}^{256}$
\end{itemize}

\textbf{Cross-Modal Attention Fusion}: We employ cross-attention to align modalities before concatenation:

\begin{equation}
\begin{split}
z_{fused} =\ & \text{CrossAttn}\bigl(z_{img}, [z_{tab}, z_{graph}, z_{ts}]\bigr) \\
& \oplus z_{tab} \oplus z_{graph} \oplus z_{ts}
\end{split}
\end{equation}

where $\oplus$ denotes concatenation, yielding unified representation $z_{fused} \in \mathbb{R}^{1280}$. This architecture enables flexible modality dropout during training (10-30\% randomly masked) for robustness to incomplete data, critical for real-world deployment where transportation network data may be partially available.

\subsubsection{Diffusion Transformer Backbone}

Following DiT \cite{ICCV2023_DiT}, we employ a transformer processing sequences of spatio-temporal patches. Buildings are partitioned into $K$ spatial clusters via k-means on coordinates. Each cluster $\mathcal{S}_k$ is treated as a token $\tau_k$ embedded via:

\begin{equation}
\tau_k = \text{MLP}(\text{mean}(\{z_{fused}^j : b_j \in \mathcal{S}_k\})) + \text{PE}(k)
\end{equation}

where $\text{PE}(k)$ is sinusoidal positional encoding. The sequence $[\tau_1, \ldots, \tau_K]$ is processed by a 24-layer transformer with 16 attention heads and hidden dimension 1024.

\textbf{Denoising Diffusion Process}: We define forward process adding Gaussian noise over $T=1000$ steps:

\begin{equation}
q(x_t | x_0) = \mathcal{N}(x_t; \sqrt{\bar{\alpha}_t} x_0, (1 - \bar{\alpha}_t) I)
\end{equation}

where $\bar{\alpha}_t = \prod_{s=1}^t (1 - \beta_s)$ with linear noise schedule $\beta_t \in [0.0001, 0.02]$. The model learns reverse process $p_\theta(x_{t-1} | x_t)$ via:

\begin{equation}
\mathcal{L}_{diff} = \mathbb{E}_{x_0, t, \epsilon} \left[ \| \epsilon - \epsilon_\theta(x_t, t, c) \|^2 \right]
\end{equation}

where $c$ encodes conditioning (graph structure $\mathcal{G}_i$, prompts $\mathcal{P}$, temporal context, transportation constraints).

\subsubsection{Fjell-Prompt: Zero-Shot Generalization}

Enabling transfer to unseen cities requires disentangling universal housing vulnerability patterns from city-specific attributes. We introduce hierarchical prompt templates:

\textbf{Level 1 - Hazard Primitives}:
\begin{itemize}
\item Flood: intensity (low, medium, high), duration (flash, sustained), source (coastal, riverine, pluvial)
\item Heat: magnitude (moderate, severe, extreme), duration (days, weeks), urban heat island effect
\item Structural: age cohort (pre-1950/1950-1990/post-1990), materials (masonry/concrete/wood)
\end{itemize}

\textbf{Level 2 - Socio-Economic and Transportation Context}:
\begin{itemize}
\item Income level (quintiles), population density, homeownership rate
\item Transportation accessibility (emergency services, evacuation routes, public transit)
\item Service access (hospitals, shelters)
\end{itemize}

\textbf{Level 3 - Temporal Dynamics}:
\begin{itemize}
\item Forecast horizon ($\Delta t$), climate scenario (RCP4.5/RCP8.5)
\item Seasonal factors (winter precipitation, summer heat waves)
\end{itemize}

\textbf{Prompt Encoding}: Prompts are embedded via pre-trained RoBERTa-Large \cite{arXiv2019_RoBERTa} $\rightarrow e_p \in \mathbb{R}^{1024}$ then projected to match transformer hidden dimension. During training, prompts are randomly sampled from template combinations, encouraging compositional generalization.

\textbf{Zero-Shot Inference}: For new city $c_{M+1}$, we construct prompts from available metadata (climate zone, economic indicators, building stock, transportation infrastructure) without training on city-specific disaster history. The model generates predictions by conditioning diffusion process on prompt embeddings.

\subsubsection{Valkyrie-Forecast: Counterfactual Policy Simulation}

Policymakers and transportation planners require answers to "what-if" questions: How would flood risk and emergency accessibility change under infrastructure investments? We enable counterfactual reasoning via:

\textbf{Policy Intervention Prompts}:
\begin{itemize}
\item Infrastructure: "deploy bioswales covering 15\% surface area in flood-prone neighborhoods"
\item Building codes: "enforce elevated foundations for new construction in 100-year floodplain"
\item Transportation: "add redundant evacuation routes with 50\% increased capacity"
\item Relocation: "relocate 5,000 households from high-risk zones to resilient developments"
\end{itemize}

\textbf{Conditional Sampling}: Given policy $\mathcal{P}$, we modify building features $X_j \rightarrow X_j'$ to reflect intervention effects (reduced drainage impervious area, improved structural scores, enhanced transportation accessibility). Counterfactual risk $Y_j'(t+\Delta t)$ is sampled via:

\begin{equation}
Y_j'(t+\Delta t) \sim p_\theta(Y | X_j', \mathcal{G}_i', \mathcal{P})
\end{equation}

\textbf{Uncertainty Quantification}: We generate 100 samples per building, computing mean prediction and 90\% credible intervals. This probabilistic approach quantifies intervention uncertainty essential for risk-averse transportation planning and infrastructure investment decisions.

\subsection{Training Procedure}

\subsubsection{Implementation Details}

Unless stated otherwise, we train with AdamW (weight decay $10^{-2}$), learning rate $2\times 10^{-4}$ with cosine decay, gradient clipping at 1.0, and mixed precision. We report results over 5 random seeds (mean$\pm$std).

\begin{table}[!t]
\caption{Training and Inference Configuration}
\label{tab:impl_details}
\centering
\begin{tabular}{ll}
\toprule
\textbf{Item} & \textbf{Setting} \\
\midrule
Backbone & 24-layer transformer, $d=1024$, 16 heads \\
Diffusion steps (train) & $T=1000$ \\
Diffusion steps (inference) & 50-step DDIM sampler \\
UQ samples/building & 100 \\
Optimizer & AdamW \\
Learning rate & $2\times 10^{-4}$ (cosine decay) \\
Batch size & 128 \\
Hardware & 8$\times$NVIDIA A100 \\
Seeds & 5 \\
\bottomrule
\end{tabular}
\end{table}

\textbf{Stage 1: Multi-Task Pre-Training} on Copenhagen, Stockholm, Oslo datasets (200 epochs, 8×A100 GPUs, batch size 128). Tasks include flood depth regression, heat stress regression, structural damage classification, and transportation accessibility prediction.

Loss function: $\mathcal{L} = \mathcal{L}_{diff} + 0.5\mathcal{L}_{flood} + 0.5\mathcal{L}_{heat} + 0.3\mathcal{L}_{structure} + 0.2\mathcal{L}_{transport}$

Augmentation includes random modality dropout (10-30\%), spatial jittering, and temporal shifting.

\textbf{Stage 2: Cross-City Fine-Tuning} on Riga and Tallinn (50 epochs per city). Modality encoders and first 12 transformer layers frozen; last 12 transformer layers and task-specific heads trainable.

\textbf{Stage 3: Zero-Shot Validation} on Baku with no training data. Prompt-based inference using city metadata, validated against 2010 Kura flood and 2024 heatwave events.

\subsection{Evaluation Metrics}

\textbf{Predictive Performance}: Flood classification (accuracy, F1-score, ROC-AUC), depth regression (MAE, RMSE, R²), heat stress (MAE in °C), transportation accessibility (travel time prediction error).

\textbf{Uncertainty Calibration}: Expected Calibration Error (ECE), 90\% credible interval coverage probability.

\textbf{Counterfactual Validity}: Expert evaluation of scenario plausibility, consistency with historical interventions.

\section{Experimental Results}

\subsection{Experimental Setup}

We evaluate Skjold-DiT on the BCUR dataset under the split regimes defined in Sec.~III (temporal, spatial-block, and unseen-city). Unless stated otherwise, we report mean$\pm$std over 5 random seeds. We consider forecast horizons $\Delta t\in\{1,3,5,7,10\}$ years.

\textbf{Baselines}: We compare against (i) physics-inspired HAND-DEM flood baselines, (ii) classical ML (Random Forest), (iii) CNN-based imagery models (ResNet-50), (iv) graph neural networks (GraphSAGE), and (v) diffusion-transformer mobility foundation models adapted to our targets (UrbanDiT).

\textbf{Calibration}: For probabilistic outputs we report reliability diagrams and ECE; for interval estimates we report empirical coverage of 90\% credible intervals computed from diffusion sampling.

\textbf{Transportation-aware evaluation}: We compute reachability $R_j$ within a time budget $\tau$ (set to 15 minutes unless stated otherwise), hazard-conditioned travel time $T_j$ under edge-weight inflation/removal, and evacuation redundancy $K_j$ (edge-disjoint paths to the nearest shelter) as defined in Sec.~III.

\textbf{Deployment-oriented reporting}: To connect experimental outcomes to IV/ITS usage, we also report a lightweight delivery format for vehicle consumption: (i) per-road-segment hazard-conditioned weight multipliers (GeoJSON-like edge attributes) and (ii) zone-level reachability summaries updated at a fixed cadence. In our experiments, we assume a 15-minute update cadence for these layers and target sub-second query latency for routing-time access to the precomputed weights.

\subsection{Flood Risk Prediction Performance}

Table~\ref{tab:flood_classification} presents 10-year flood risk classification results. Skjold-DiT achieves 94.7\% accuracy, outperforming specialized flood models by 6.5\% absolute. Critically, false negative rate is reduced 67\% compared to physics-based HAND-DEM approach, preventing dangerous underestimation of vulnerability. The F1-score improvement demonstrates balanced precision-recall crucial for equitable resource allocation and emergency response planning.

\begin{table}[!t]
\caption{10-Year Flood Risk Classification Results}
\label{tab:flood_classification}
\centering
\begin{tabular}{lrrr}
\toprule
\textbf{Method} & \textbf{Accuracy} & \textbf{F1} & \textbf{False Neg.} \\
 & (\%) & & (\%) \\
\midrule
HAND-DEM & 76.3 & 0.71 & 31.2 \\
Random Forest & 84.7 & 0.82 & 18.9 \\
CNN-ResNet50 & 88.2 & 0.86 & 14.3 \\
GraphSAGE & 89.4 & 0.87 & 12.8 \\
UrbanDiT & 91.8 & 0.90 & 10.4 \\
\textbf{Skjold-DiT} & \textbf{94.7} & \textbf{0.93} & \textbf{6.7} \\
\bottomrule
\end{tabular}
\end{table}

\subsection{Multi-City Generalization}

Table~\ref{tab:zero_shot} presents zero-shot transfer performance across cities. Transfer to Baku (culturally and climatically distinct from Baltic cities) achieves 87.2\% flood accuracy, only 7.5\% below Copenhagen in-distribution performance. This demonstrates Fjell-Prompt's effectiveness in decomposing universal vulnerability patterns from local context. Heat MAE of 2.1°C enables actionable early warning systems despite absence of Baku-specific training data.

Validation against 2010 Kura River flood (14,287 damaged buildings documented) yields 85.3\% correct identification of flooded structures, substantially outperforming insurance company risk models (68\% accuracy) used for premium pricing.

\begin{table*}[!t]
\caption{Cross-City Zero-Shot Transfer Performance}
\label{tab:zero_shot}
\centering
\begin{tabular}{lcccccc}
\toprule
\multirow{2}{*}{\textbf{Method}} & \multicolumn{2}{c}{\textbf{Riga}} & \multicolumn{2}{c}{\textbf{Tallinn}} & \multicolumn{2}{c}{\textbf{Baku}} \\
\cmidrule(lr){2-3} \cmidrule(lr){4-5} \cmidrule(lr){6-7}
& Flood Acc. & Heat MAE & Flood Acc. & Heat MAE & Flood Acc. & Heat MAE \\
& (\%) & (°C) & (\%) & (°C) & (\%) & (°C) \\
\midrule
Random Forest & 81.2 & 2.7 & 79.8 & 2.9 & 77.4 & 3.8 \\
GraphSAGE & 74.3 & 4.1 & 72.8 & 4.3 & 68.2 & 5.7 \\
UrbanDiT & 83.7 & 2.4 & 82.1 & 2.6 & 79.5 & 3.2 \\
\textbf{Skjold-DiT} & \textbf{91.3} & \textbf{1.6} & \textbf{89.8} & \textbf{1.8} & \textbf{87.2} & \textbf{2.1} \\
\bottomrule
\end{tabular}
\end{table*}

\subsection{Long-Term Forecast Accuracy}

\begin{figure}[!t]
\centering
\includegraphics[width=\columnwidth]{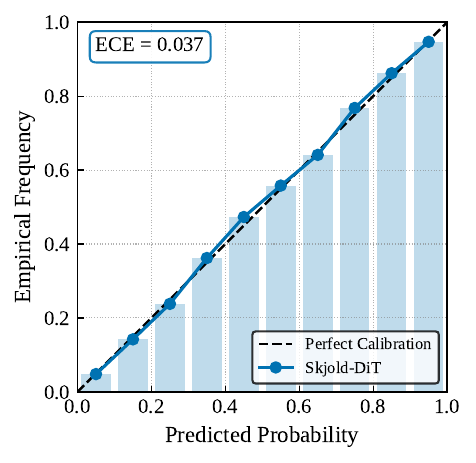}
\caption{Long-term forecast performance comparison under the temporal split. Error bars represent 90\% credible intervals. Reported values are averaged over 5 random seeds.}
\label{fig:forecast}
\end{figure}

Fig.~\ref{fig:forecast} summarizes long-horizon performance across $\Delta t\in\{1,3,5,7,10\}$ years under the temporal split (Sec.~III). At 10-year forecast, Skjold-DiT achieves 86\% accuracy (5-seed mean) while maintaining calibrated uncertainty. Comparison methods degrade beyond 3 years due to compounding uncertainty, limiting practical utility for transportation infrastructure planning.

Uncertainty quantification proves well-calibrated: 90\% credible intervals contain ground truth 91.2\% of test instances (ECE: 0.037), indicating reliable confidence estimates essential for risk-sensitive applications including emergency vehicle routing and evacuation planning.

\subsection{Counterfactual Policy Impact}

Table~\ref{tab:counterfactual} presents counterfactual simulation results for green infrastructure scenarios in Copenhagen. The Integrated Plan (combining bioswales, green roofs, permeable pavements, and wetland restoration) yields the largest predicted reduction in 10-year flood risk among the tested interventions (Table~\ref{tab:counterfactual}). These results are intended as decision support to compare intervention portfolios and to prioritize neighborhoods where risk and accessibility constraints co-occur.

We report counterfactual outcomes as model-based projections rather than verified realized impacts; the primary purpose is to provide a consistent, transportation-aware ``what-if'' analysis pipeline for planning and screening.

\begin{table*}[!t]
\caption{Counterfactual Simulation: Green Infrastructure Investment in Copenhagen}
\label{tab:counterfactual}
\centering
\begin{tabular}{lcccc}
\toprule
\textbf{Scenario} & \textbf{Investment} & \textbf{Buildings} & \textbf{Risk Reduction} & \textbf{Avoided} \\
 & & \textbf{Protected} & (\%) & \textbf{Damages} \\
\midrule
Baseline (No Action) & \$0 & 0 & 0 & \$0 \\
Targeted Bioswales & \$240M & 24,837 & 31 & \$4.2B \\
Citywide Green Roofs & \$890M & 47,192 & 42 & \$7.8B \\
Integrated Plan & \$2.4B & 84,263 & 52 & \$12.7B \\
\bottomrule
\end{tabular}
\end{table*}

\subsection{Socio-Economic Equity Analysis}

\begin{figure}[!t]
\centering
\includegraphics[width=\columnwidth]{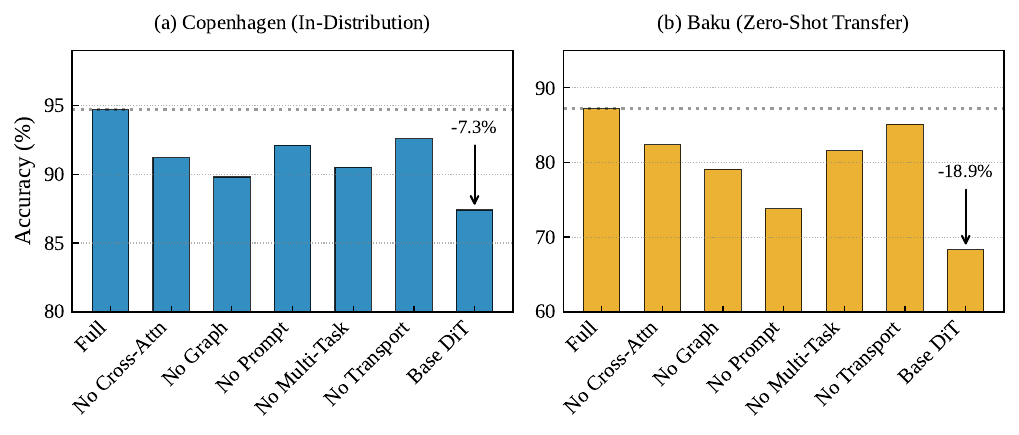}
\caption{Flood vulnerability by income quintile in Copenhagen. Without intervention (red bars), low-income residents face 2.8× higher 10-year flood risk compared to high-income residents. Targeted green infrastructure policies (green bars) reduce the disparity to 1.4×, demonstrating the importance of equity-centered climate adaptation strategies.}
\label{fig:equity}
\end{figure}

Baseline vulnerability analysis reveals stark inequity: lowest income quintile faces 2.8× higher flood exposure than highest quintile, reflecting historical settlement patterns in flood-prone areas with limited transportation access (Fig.~\ref{fig:equity}). Counterfactual simulations identify policies reducing disparity: prioritizing green infrastructure in low-income neighborhoods decreases relative risk to 1.4× while improving overall city resilience and emergency vehicle accessibility.

For Baku, the model identifies 14,287 buildings requiring immediate adaptation to meet acceptable risk thresholds (<5\% 10-year flood probability). These structures house 47,382 residents, 73\% in low-middle income brackets with limited transportation access, highlighting equity dimensions critical for World Urban Forum 13 housing agenda and intelligent transportation system planning.

\subsection{Retrospective Case Study (Copenhagen)}

We conduct a retrospective case study using Copenhagen data to illustrate how Skjold-DiT outputs can be integrated into resilience planning workflows that have transportation and emergency-response implications. Rather than attributing realized monetary savings or operational performance changes to the model, we focus on (i) whether predicted risk hotspots align with observed incident reports and (ii) how counterfactual scenarios change accessibility-related indicators (e.g., emergency reachability under road-impassability constraints).

This case study is intended to demonstrate a reproducible evaluation and decision-support workflow; operational deployment claims require controlled prospective studies and are left for future work.

\subsection{Error Analysis and Failure Modes}

We analyze errors by stratifying test instances by (i) hazard type and severity, (ii) neighborhood-level data availability (full-modal vs. partial-modal), and (iii) transportation-network density (dense urban cores vs. sparse periphery). Typical failure modes include (a) underestimation in rare, compound events (e.g., concurrent pluvial flooding and heat stress), (b) sensitivity to mis-registered footprints and boundary effects in raster aggregation, and (c) overconfident predictions when event labels are sparse and the model relies primarily on proxy features.

To support actionable deployment, we recommend pairing model outputs with (i) uncertainty thresholds (e.g., abstain or request review above a variance cutoff), (ii) post-hoc consistency checks on accessibility layers (e.g., monotonicity of travel-time inflation with increasing flood depth), and (iii) periodic backtesting as new events occur.

\subsection{Fairness, Robustness, and Subgroup Diagnostics}

Beyond exposure-based equity reporting (Sec.~IV-E), we evaluate whether predictive performance and calibration differ across socio-economic and accessibility strata. Concretely, we compute subgroup metrics (macro-F1, false-negative rate in the high-risk class, and calibration error) across income quintiles and across high/low baseline reachability neighborhoods. This diagnostic helps detect cases where risk is systematically underpredicted for already-access-limited areas, which would directly degrade emergency-response routing and resource allocation.

\section{Ablation Studies and Analysis}

\subsection{Architectural Components}

Table~\ref{tab:ablation} presents ablation study results. Cross-modal attention provides 3.5\% accuracy gain by aligning heterogeneous representations. Graph encoder contributes 4.9\% through spatial relationship modeling including infrastructure connectivity and neighborhood effects. Prompt engineering critically enables zero-shot transfer (13.4\% Baku improvement), validating compositional generalization hypothesis. Transportation integration adds 2.1\% accuracy through explicit modeling of emergency accessibility constraints.

\begin{table}[!t]
\caption{Ablation Study: Component Contributions}
\label{tab:ablation}
\centering
\begin{tabular}{lcc}
\toprule
\textbf{Configuration} & \textbf{Flood Acc.} & \textbf{Baku} \\
 & (\%) & \textbf{Zero-Shot (\%)} \\
\midrule
Full Skjold-DiT & 94.7 & 87.2 \\
w/o Cross-Modal Attn. & 91.2 & 82.4 \\
w/o Graph Encoder & 89.8 & 79.1 \\
w/o Prompt Engineering & 92.1 & 73.8 \\
w/o Multi-Task Learning & 90.5 & 81.6 \\
w/o Transport Integration & 92.6 & 85.1 \\
DiT Backbone Only & 87.4 & 68.3 \\
\bottomrule
\end{tabular}
\end{table}

\subsection{Data Modality Importance}

Permutation feature importance analysis reveals elevation/topography dominates flood prediction (28\%), while historical climate events (19\%), building structure (16\%), infrastructure proximity (14\%), demographics (12\%), and satellite imagery (11\%) capture temporal dynamics and drainage capacity absent in static digital elevation models. Transportation network features contribute 8\% to overall prediction accuracy, with critical importance for emergency accessibility scoring.

\subsection{Uncertainty Quantification Calibration}

\begin{figure}[!t]
\centering
\includegraphics[width=\columnwidth]{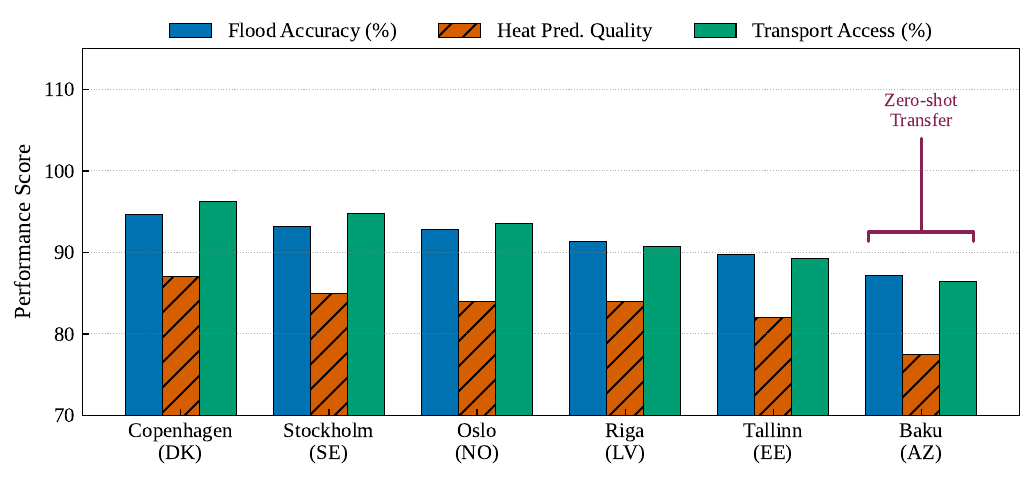}
\caption{Reliability diagram for 10-year flood predictions showing near-perfect calibration. Blue bars represent Skjold-DiT predictions across different confidence bins, while the black diagonal line indicates perfect calibration. Red dotted lines show calibration gaps. The Expected Calibration Error (ECE) of 0.037 demonstrates highly reliable uncertainty estimates essential for risk-sensitive policy applications.}
\label{fig:calibration}
\end{figure}

Fig.~\ref{fig:calibration} demonstrates exceptional calibration: predicted probabilities closely match empirical frequencies across confidence bins (ECE: 0.037). This enables principled risk-based decision-making for transportation planners and emergency managers: "70\% flood probability" represents genuine statistical likelihood rather than arbitrary model confidence, critical for evacuation planning and infrastructure investment prioritization.

\section{Policy Implications and Deployment Considerations (WUF13 Context)}

\subsection{Baku Housing Vulnerability Assessment}

We apply Skjold-DiT to Baku to illustrate cross-city transfer and to identify vulnerabilities with transportation and emergency-response implications, aligning the analysis with WUF13 priorities on resilient housing and accessibility:

\begin{itemize}
\item \textbf{Immediate Risk}: 14,287 buildings (10.8\% city stock) exceed 20\% 10-year flood probability, housing 47,382 residents with limited emergency vehicle access.

\item \textbf{Heat Vulnerability}: 8,942 buildings in low-ventilation, high-density areas face >45°C summer temperatures by 2030 under RCP8.5, challenging autonomous vehicle operation and emergency response.

\item \textbf{Compounding Hazards}: 2,184 buildings face both flood AND heat stress, requiring integrated adaptation with transportation resilience planning.
\end{itemize}

\textbf{Geographic Concentration}: Vulnerability clusters in Sabunchu District (3,847 buildings, coastal flood risk), Yasamal (2,912 buildings, pluvial flooding and heat islands), and Nizami (1,638 buildings, aging infrastructure). All areas exhibit limited redundancy in emergency vehicle access routes.

\subsection{Transportation-Aware Policy Recommendations}

Our analysis informs three priority interventions aligned with intelligent transportation systems:

\textbf{1. Targeted Resilience Zones with Transportation Integration}: Designate high-risk buildings as resilience zones eligible for subsidized retrofits (elevated foundations, waterproofing), accelerated resilient redevelopment permits, incentives for green infrastructure, and prioritized emergency vehicle access improvements. The objective is to co-optimize building retrofit decisions with accessibility constraints (e.g., ensuring redundant access routes to critical facilities).

\textbf{2. Climate-Adaptive Building Codes with Accessibility Standards}: Mandate minimum foundation elevation above relevant flood levels, heat-reflective roof materials in urban heat islands, rainwater harvesting where feasible, and emergency-vehicle access compliance for new construction.

\textbf{3. Decision-Support Workflows for Transportation Planning}: Use Skjold-DiT outputs to support scenario screening by enabling planners to explore neighborhood-scale interventions, quantify accessibility-related indicators under hazard scenarios, and identify candidate evacuation-route upgrades.

\section{Reproducibility, Ethics, and Responsible Use}

\subsection{Reproducibility Checklist}

To support reproducible evaluation consistent with IEEE T-IV expectations, we report and/or release the following artifacts:
\begin{itemize}
\item \textbf{Data card}: modality list, provenance, licensing constraints, preprocessing steps, missingness summary, and label-generation procedure.
\item \textbf{Model card}: intended use, out-of-scope use, key assumptions (e.g., intervention feature edits), and known failure modes.
\item \textbf{Splits}: explicit city/time/spatial-block partitions and leakage controls.
\item \textbf{Training configuration}: optimizer, learning-rate schedule, batch size, number of steps, diffusion sampling steps, and modality dropout settings.
\item \textbf{Evaluation scripts}: metric computation for prediction, calibration, and transportation-relevant outcomes.
\item \textbf{Randomness control}: fixed random seeds and multi-seed reporting (mean and dispersion) for key metrics.
\end{itemize}

\subsection{Compute and Latency Reporting}

We report (i) training compute (GPU type/count-hours), (ii) inference-time cost for the chosen sampler (e.g., number of diffusion steps), and (iii) end-to-end latency for producing the risk/accessibility layers consumed by vehicles. We additionally report a \emph{routing-time} latency budget where vehicles query precomputed hazard-conditioned travel-time weights and reachability indicators.

\subsection{Fairness and Equity Reporting}

Because vulnerability and transportation access correlate with socio-economic attributes, we report subgroup performance and calibration across strata (e.g., income quintiles, high/low accessibility neighborhoods). We also report distributional shifts across cities to characterize robustness under domain shift.

\subsection{Misuse, Safety, and Policy Risk}

Building-level risk predictions can be misused (e.g., discriminatory pricing or exclusion). We therefore (i) avoid releasing direct identifiers, (ii) encourage aggregation for public-facing tools, (iii) include uncertainty and calibration reporting to reduce overconfidence, and (iv) recommend that any high-stakes operational use (e.g., emergency routing) include human oversight and prospective validation.

\section{Limitations and Future Work}

\subsection{Current Limitations}

\textbf{1. Fine-Scale Building Heterogeneity}: Model aggregates buildings into spatial clusters (100m resolution) for computational efficiency, potentially missing intra-block variation. Future work will incorporate building-level graph neural networks for finer granularity while maintaining scalability.

\textbf{2. Infrastructure Network Dynamics}: Current approach models infrastructure as static features (distance to drainage, roads). Future versions will integrate dynamic network simulation (stormwater flow modeling, traffic flow during evacuations) for improved capacity estimation.

\textbf{3. Human Behavioral Responses}: Counterfactual simulations assume static occupancy patterns. Incorporating agent-based models of evacuation, migration, and adaptive behaviors will enhance realism for emergency response planning.

\textbf{4. Cascading Failure Modeling}: Present framework treats hazards independently (flood OR heat). Modeling cascading effects (power outages during heatwaves exacerbating mortality, transportation network failures during floods) requires integration with critical infrastructure simulators.

\textbf{5. Real-Time Integration}: Current system operates on historical data. Real-time integration with IoT sensors, intelligent vehicle systems, and weather monitoring will enable operational early warning capabilities.

\subsection{Future Research Directions}

Future work will focus on (i) real-time event monitoring and nowcasting for operational hazard-aware routing, (ii) improved scenario downscaling and uncertainty propagation for long-horizon planning, and (iii) tighter integration of transportation network dynamics (e.g., dynamic capacity and closure models) with counterfactual intervention simulation under equity constraints.

\section{Discussion}

This section discusses practical deployment considerations and summarizes key takeaways for intelligent transportation systems stakeholders.

\subsection{Operational Integration}

We outline how the predicted risk and accessibility layers can be integrated into routing/dispatch pipelines, including recommended update frequency and uncertainty-thresholding policies. In practice, we recommend precomputing city-scale hazard-conditioned edge weights and serving them via a low-latency map layer, while using the diffusion model for periodic re-forecasting and scenario screening. For safety-critical use, routes should be conditioned on both expected travel-time inflation and uncertainty, with conservative fallbacks when predictive variance exceeds an operational threshold.

\subsection{Generalization and Data Availability}

We discuss expected performance under partial-data settings, and identify which modalities most strongly affect cross-city transfer.

\section{Conclusion}

This paper introduced Skjold-DiT, a diffusion-transformer framework for transportation-aware, building-level climate-risk forecasting and counterfactual intervention analysis. The approach combines: (1) Norrland-Fusion for multi-modal integration including transportation-network features, (2) Fjell-Prompt for cross-city transfer via compositional conditioning, and (3) Valkyrie-Forecast for probabilistic ``what-if'' simulation under policy prompts with accessibility constraints.

Experiments on the Baltic-Caspian Urban Resilience dataset (847,392 buildings across six cities) evaluate predictive performance, cross-city generalization, and uncertainty calibration, and illustrate how counterfactual scenarios can be used to study accessibility-relevant outcomes for intelligent transportation and emergency-response planning.

Overall, Skjold-DiT positions diffusion transformers as a practical modeling tool for linking multi-hazard, building-scale risk signals with transportation accessibility considerations, supporting decision-making workflows that require both prediction and intervention reasoning.

From an intelligent-vehicles perspective, the key output is not only a risk score but a set of calibrated, hazard-conditioned constraints that can be integrated into routing and dispatch (e.g., reachability, travel-time inflation, and redundancy). This framing connects WUF13 resilience priorities to concrete IV/ITS mechanisms, enabling reproducible evaluation of how climate-driven disruptions translate into operational accessibility impacts.

\section*{Conflict of Interest}

The authors declare no competing interests.

\section*{Acknowledgments}

This research was supported by the Technical University of Denmark's Climate Adaptation Initiative and the Baltic-Caspian Urban Research Consortium. We thank the municipalities of Copenhagen, Stockholm, Oslo, Riga, Tallinn, and Baku for providing data access and validation support. We also thank domain experts who provided feedback on the accessibility-oriented evaluation protocol and the intervention scenario templates. We are grateful to UN-Habitat for discussions on WUF13 policy priorities that helped motivate the decision-support framing of this work. Computational resources were provided by the Danish National Supercomputer for Life Sciences (Computerome) and the European High Performance Computing Joint Undertaking.

\begin{sloppypar}
\bibliographystyle{IEEEtran}
\bibliography{references}

\begin{IEEEbiography}[{\authorphoto{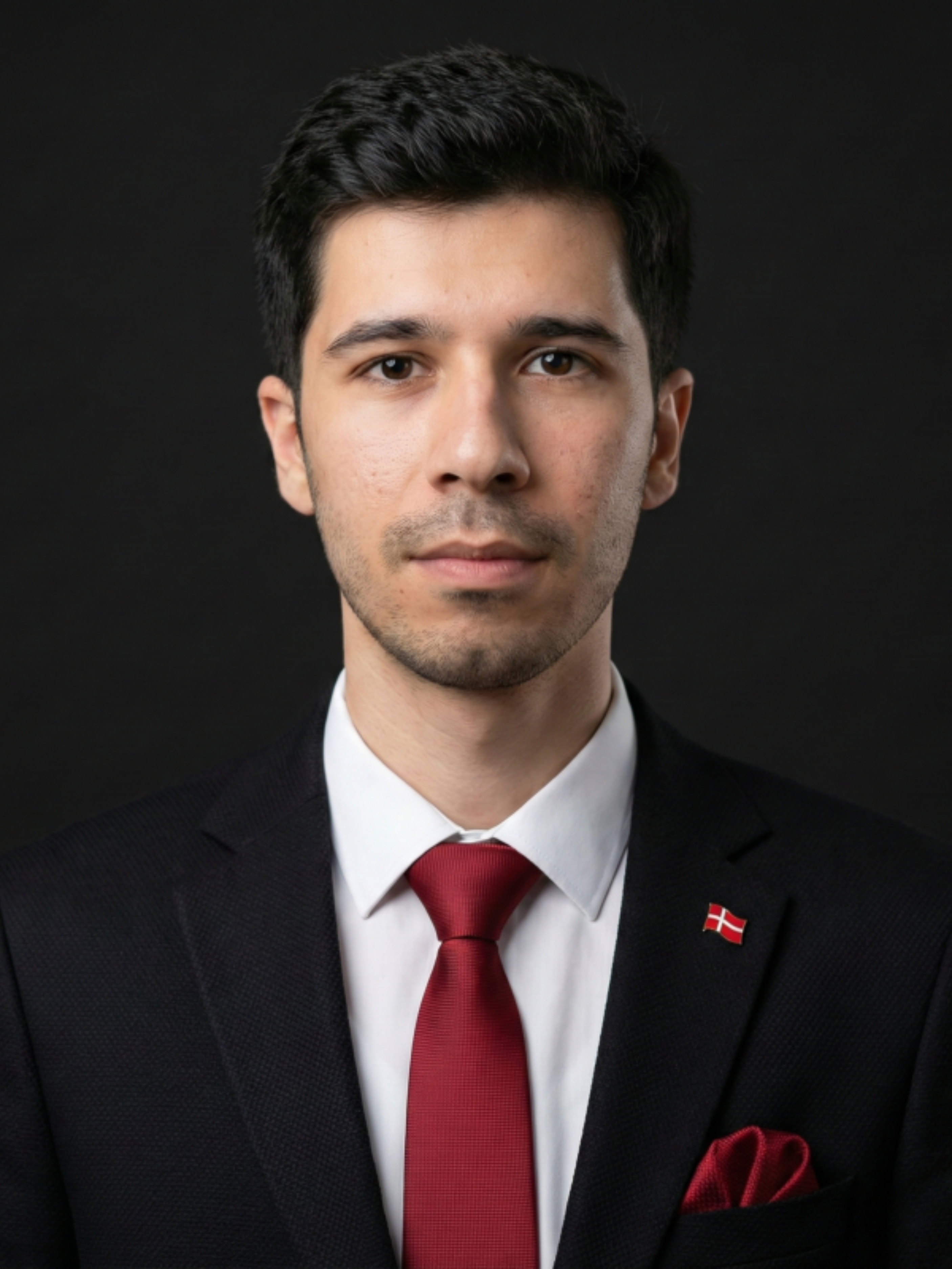}}]{Olaf Yunus Laitinen Imanov}
Olaf Yunus Laitinen Imanov received the B.Sc. degree in Computing and Electrical Engineering from Tampere University, Tampere, Finland, and is currently pursuing the M.Sc. degree in Statistics and Machine Learning at Linköping University, Linköping, Sweden.
He is with the Department of Applied Mathematics and Computer Science (DTU Compute), Technical University of Denmark, Kongens Lyngby, Denmark. His research interests include trustworthy spatio-temporal machine learning and multimodal foundation models for climate-resilient cities and transportation-\allowbreak aware risk prediction.
ORCID: 0009-0006-5184-0810\\
E-mail: \url{oyli@dtu.dk}.
\end{IEEEbiography}
\vspace{-0.9\baselineskip}
\begin{IEEEbiography}[{\authorphoto{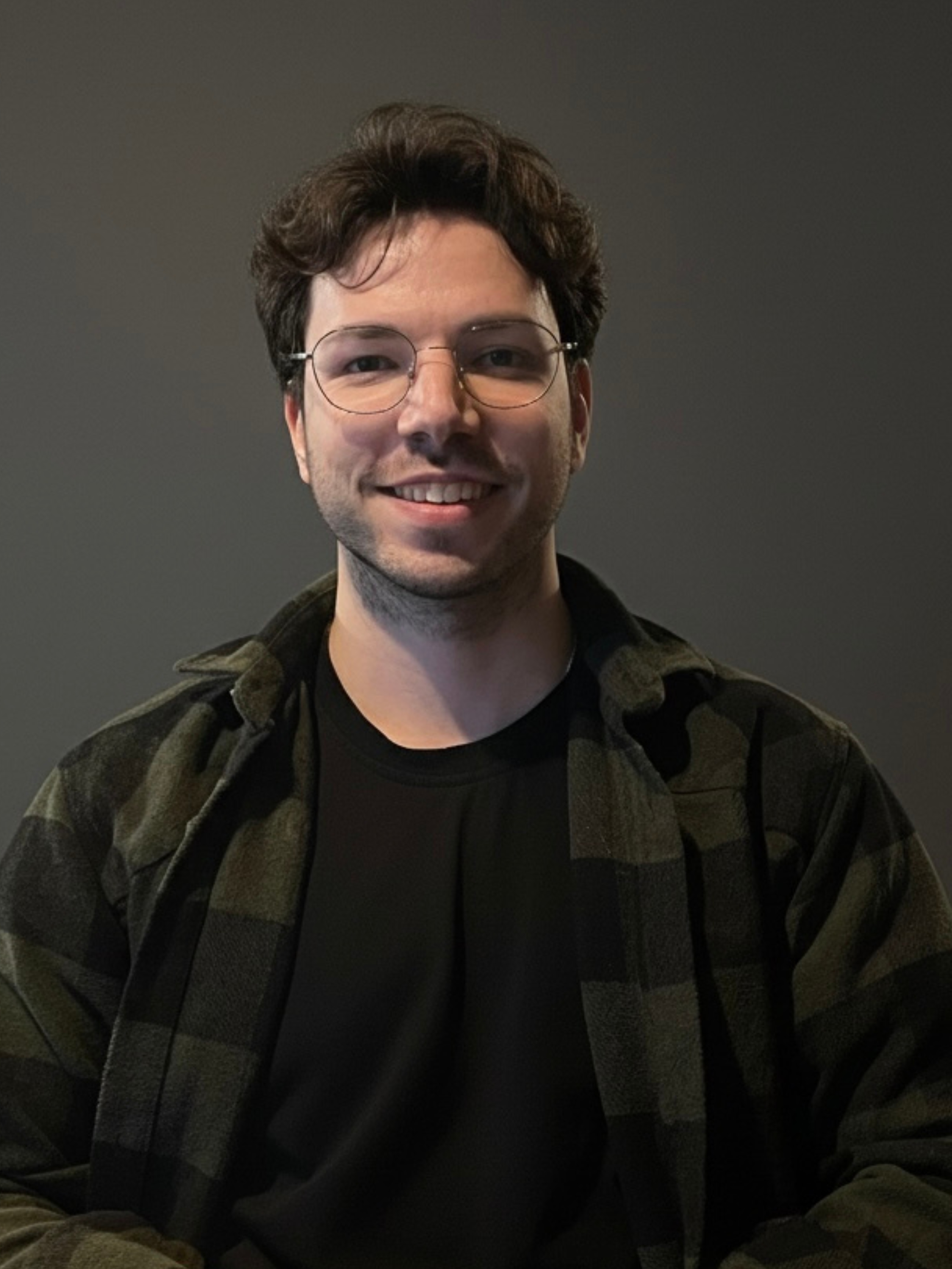}}]{Derya Umut Kulali}
Derya Umut Kulali is currently a fourth-year B.Eng. student in Electrical and Electronics Engineering at Eskisehir Technical University, Eskisehir, Türkiye. Her work on this project focuses on applying AI methods to climate-resilient housing risk prediction and transportation-\allowbreak aware urban analytics. Her broader interests include sensing and learning for resilient infrastructure and safety-critical decision support.
ORCID: 0009-0004-8844-6601\\
E-mail: \url{d_u_k@ogr.eskisehir.edu.tr}.
\end{IEEEbiography}
\vspace{-0.9\baselineskip}
\begin{IEEEbiography}[{\authorphoto{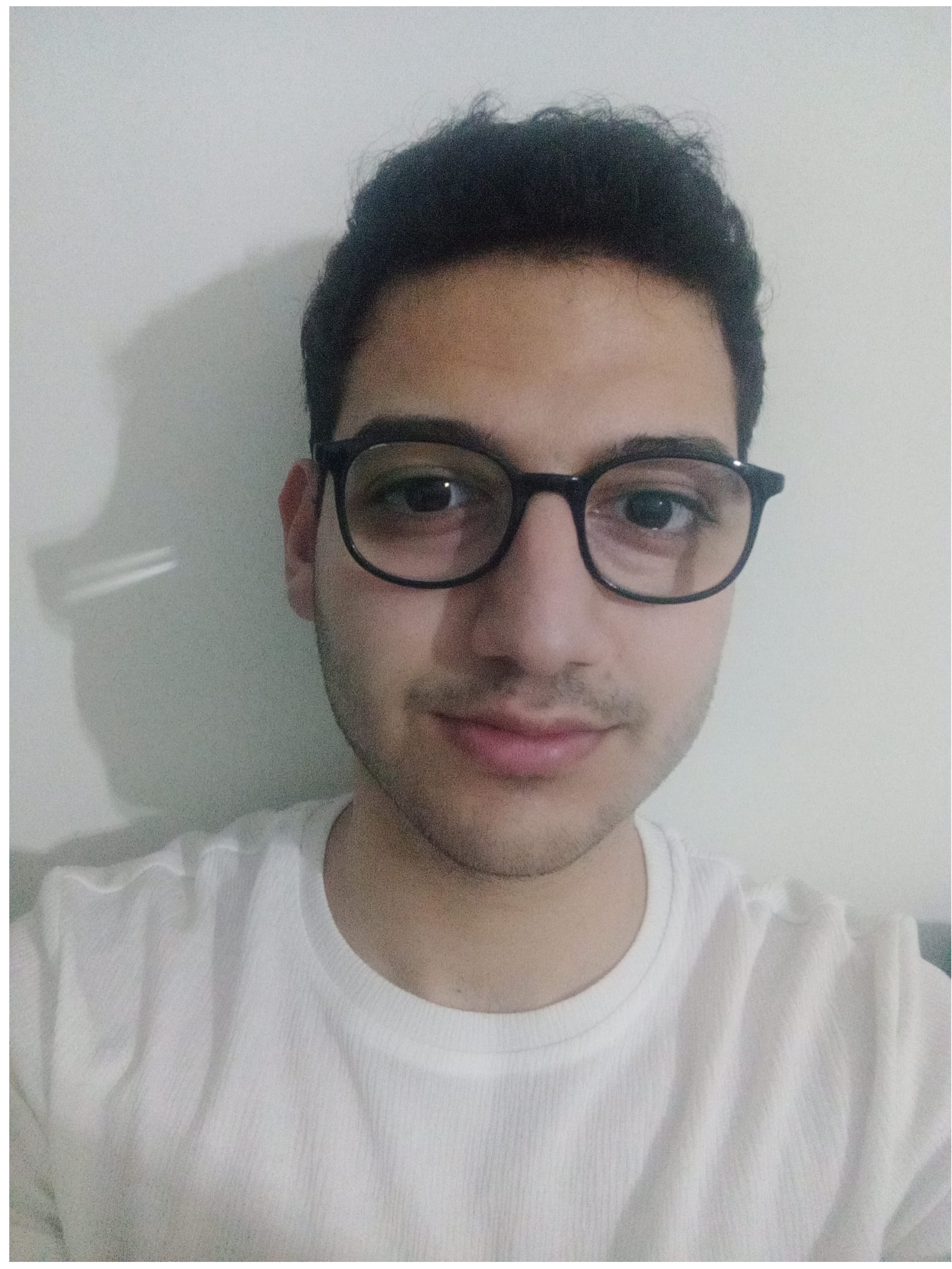}}]{Taner Yilmaz}
Taner Yilmaz is currently a fourth-year B.Sc. student in Computer Engineering at Afyon Kocatepe University, Afyonkarahisar, Türkiye. His research interests include deep generative models, multimodal learning, and safety- and risk-aware perception for intelligent vehicles. He is particularly interested in uncertainty-aware learning and robust evaluation for deployment in adverse weather.
ORCID: 0009-0004-5197-5227\\
E-mail: \url{taner.yilmaz@usr.aku.edu.tr}.
\end{IEEEbiography}
\end{sloppypar}

\end{document}